\documentclass[11pt, oneside, round]{article}   	
\usepackage{geometry}                		
\usepackage{amsmath}
\usepackage{amssymb}
\usepackage{amsthm}
\usepackage{anysize}
\usepackage{authblk}
\usepackage{bm}
\usepackage{booktabs}
\usepackage{cancel}
\usepackage{enumitem}
\usepackage{float}
\usepackage{mathtools}
\usepackage{multirow}
\usepackage{natbib}
\usepackage{times}
\usepackage[colorlinks=true,linkcolor=red,citecolor=blue,urlcolor=blue,pagebackref]{hyperref}
\usepackage[linesnumbered,ruled]{algorithm2e}
\usepackage{lineno}
\usepackage{pdflscape}
\usepackage{stmaryrd}
\usepackage{adjustbox}
\usepackage{tikz}
\usepackage{neuralnetwork}

\SetCommentSty{mycommfont}

\usetikzlibrary{arrows.meta} 
\usepackage[outline]{contour} 
\contourlength{1.4pt}

\tikzset{>=latex} 
\usepackage{xcolor}
\colorlet{myred}{red!80!black}
\colorlet{myblue}{blue!80!black}
\colorlet{mygreen}{green!60!black}
\colorlet{myorange}{orange!70!red!60!black}
\colorlet{mydarkred}{red!30!black}
\colorlet{mydarkblue}{blue!40!black}
\colorlet{mydarkgreen}{green!30!black}
\tikzstyle{node}=[thick,circle,draw=myblue,minimum size=22,inner sep=0.5,outer sep=0.6]
\tikzstyle{node in}=[node,green!20!black,draw=mygreen!30!black,fill=mygreen!25]
\tikzstyle{node hidden}=[node,blue!20!black,draw=myblue!30!black,fill=myblue!20]
\tikzstyle{node convol}=[node,orange!20!black,draw=myorange!30!black,fill=myorange!20]
\tikzstyle{node out}=[node,red!20!black,draw=myred!30!black,fill=myred!20]
\tikzstyle{connect}=[thick,mydarkblue] 
\tikzstyle{conred}=[thick,mydarkred]
\tikzstyle{connect arrow}=[-{Latex[length=4,width=3.5]},thick,mydarkblue,shorten <=0.5,shorten >=1]
\tikzset{ 
  node 1/.style={node in},
  node 2/.style={node hidden},
  node 3/.style={node out},
}

\DeclareMathOperator{\Hess}{Hess}
\DeclareMathOperator{\Var}{Var}
\DeclareMathOperator{\trace}{trace}
\DeclareMathOperator{\Span}{span}

\DeclareMathOperator{\train}{train}
\DeclareMathOperator{\val}{val}

\DeclareMathOperator{\LSMR}{LSMR}

\DeclareMathOperator{\sqr}{sqrt}
\DeclareMathOperator{\zeros}{zeros}

\DeclareMathOperator{\avg}{avg}
\DeclareMathOperator{\rel}{rel}
\DeclareMathOperator{\maxiter}{maxiter}
\DeclareMathOperator{\drop}{drop}
\DeclareMathOperator{\atol}{atol}
\DeclareMathOperator{\ftol}{ftol}

\makeatletter
\newcommand{\longdash}[1][2em]{%
  \makebox[#1]{$\m@th\smash-\mkern-7mu\cleaders\hbox{$\mkern-2mu\smash-\mkern-2mu$}\hfill\mkern-7mu\smash-$}}
\makeatother
\newcommand{\omitskip}{\kern-\arraycolsep}
\newcommand{\llongdash}[1][1.5em]{\longdash[#1]\omitskip}
\newcommand{\rlongdash}[1][1.5em]{\omitskip\longdash[#1]}

\newcommand{\dd}{\mathrel{\!:\!}}
\providecommand{\keywords}[1]{\textbf{Key words:} #1}
\providecommand{\AMS}[1]{\textbf{AMS subject classifications:} #1}

\newcommand{\R}{\mathbb{R}}

\newcommand{\E}{\mathbb{E}}
\newcommand{\N}{\mathbb{N}}
\newcommand{\W}{\mathcal{W}}
\newcommand{\X}{\mathcal{X}}

\graphicspath{{../figures/}} 

\title{Training Autoencoders Using Stochastic Hessian-Free Optimization with LSMR}

\author[1]{Ibrahim Emirahmetoglu}
\author[1]{David E.\,Stewart}
\affil[1]{Department of Mathematics, University of Iowa}

\begin{document}
\maketitle

\begin{abstract}
\noindent
Hessian-free (HF) optimization has been shown to effectively train deep autoencoders \citep{martens2010deep}. In this paper, we aim to accelerate HF training of autoencoders by reducing the amount of data used in training. HF utilizes the conjugate gradient algorithm to estimate update directions. Instead, we propose using the LSMR method, which is known for
effectively solving large sparse linear systems. We also incorporate \citet{chapelle2011improved}'s improved preconditioner for HF optimization. In addition, we introduce a new mini-batch selection algorithm to mitigate overfitting. Our algorithm starts with a small subset of the training data and gradually increases the mini-batch size based on (i) variance estimates obtained during the computation of a mini-batch gradient \citep{byrd2012sample} and (ii) the relative decrease in objective value for the validation data. Our experimental results demonstrate that our stochastic Hessian-free optimization, using the LSMR method and the new sample selection algorithm, leads to rapid training of deep autoencoders with improved generalization error.
\end{abstract}


\vspace{2mm}
\AMS{Primary: 65K05, Secondary: 49M15}

\section{Introduction}
Second-order optimization techniques have been widely studied for problems suffering from pathological curvature, such as training deep neural networks. In particular, the second-order method that does not rely on computing the full Hessian matrix, known as Hessian-free (HF) optimization, has been shown to effectively train deep autoencoders \citep{martens2010deep}.

\vspace{2mm}\noindent
The primary challenges of HF optimization are slow training and overfitting. In this study, we aim to speed up HF training of autoencoders by reducing the amount of data used in training. The bottleneck of HF optimization is the need to solve large linear systems to estimate update directions. Many papers in the literature on HF optimization use the conjugate gradient (CG) algorithm to solve these large linear systems. In contrast, we propose using the Least Squares Minimal Residual method \citep[LSMR;][]{fong2011lsmr}, which is known for effectively solving large sparse linear systems. In practice, LSMR often converges faster than CG for least-squares problems, which is beneficial when training deep autoencoders on large datasets. Since both the residuals for iterates and the residuals of normal equations decrease monotonically, it is safer to terminate LSMR early \citep{fong2011lsmr}, which reduces the number of Krylov subspace iterations required for a solution to approximate the Hessian within each HF iteration.

\vspace{2mm}\noindent
Our second contribution is to introduce a new mini-batch selection algorithm, which helps mitigate overfitting and reduce the amount of data used in training. Our algorithm begins with a small subset of the training data, known as a mini-batch, and then progressively increases the mini-batch size based on (i) variance estimates acquired during the computation of a mini-batch gradient \citep*{byrd2012sample} and (ii) a relative decrease in the objective value for the validation data. Note that using a larger sample to compute the gradients than the curvature matrix-vector products may improve the quality of the updates, but it can also have negative effects \citep{martens2012training}. Our mini-batch selection algorithm allows us to use the same sample to compute the gradient and Hessian.

\vspace{2mm}\noindent
The updates made by HF may be overly tailored to the current mini-batch of training data, leading to overfitting. Our algorithm avoids overfitting in three ways. First, unlike \citet{martens2010deep}, we do not cycle through mini-batches during epochs. Due to the stochastic nature of our algorithm, we never use the same mini-batch in HF iterations. Second, increasing the size of the mini-batches allows LSMR to obtain more accurate estimates of the gradient and curvature. This aligns with the observation by \citet{martens2012training} that the mini-batch overfitting problem worsens as optimization progresses in training neural networks. Therefore, continually increasing the mini-batch size can help mitigate overfitting. Third, monitoring the relative decrease in objective value for the validation data and using it as a termination condition for LSMR is a highly effective way to prevent mini-batch overfitting.

\vspace{2mm}\noindent
Another contribution of this work is to eliminate the need for converting weight matrices into vectors. In deep neural networks, weight matrices are typically vectorized so that the parameters of interest can be written as a vector. However, in this paper, we represent the weights as a vector composed of matrices, which helps us avoid multiple matrix-vector conversions. This approach enables us to efficiently compute gradients using matrix algebra tools in Julia \citep{bezanson2012julia}. The details are provided in Section \ref{notation}.

\vspace{2mm}\noindent
The remainder of the paper is as follows: In Section \ref{related}, we discuss related works of HF optimization for training deep neural networks. In Section \ref{autoencoders}, we introduce autoencoders and our notation. Section \ref{SHF} investigates Hessian-free optimization and presents our contribution to HF optimization in a stochastic setting. In Section \ref{LSMR}, the LSMR method is introduced for solving sparse least-squares problems. Preliminary experiments to demonstrate the performance of our stochastic HF algorithm are in Section \ref{experiments}. Finally, Section \ref{conclusion} concludes this work.

\section{Related works} \label{related}
In this section, we present a brief summary of the main works that apply Hessian-free optimization to deep neural networks. In a seminal work, \citet{martens2010deep} successfully applies Hessian-free optimization for training autoencoders. Following \citet{martens2010deep} and introducing a novel damping scheme, \citet{martens2011learning} effectively train recurrent neural networks on complex and challenging sequence modeling problems. \citet*{martens2012training} discuss various practical Hessian-free techniques for feed-forward and recurrent neural networks, including damping mechanisms, preconditioning, and analysis of mini-batch gradient and Hessian information. \citet{kingsbury2012scalable} describe a distributed neural network training algorithm, based on HF optimization, that scales to deep networks and large datasets.

\vspace{2mm}\noindent
To train autoencoders and fully-connected neural networks, \citet{kiros2013training} considers stochastic HF with gradient and curvature mini-batches independent of the dataset size and integrates dropout to guard against overfitting. Throughout the training, a constant gradient batch size of $1000$, curvature batch size of $100$, and 3 CG iterations per batch are used. Another paper related to ours is \citet{sainath2013accelerating}, which accelerates HF optimization for deep neural networks by developing an L-BFGS based preconditioning scheme that avoids the need to access the Hessian explicitly and proposing a new sampling algorithm. 

\vspace{2mm}\noindent
\citet*{wang2015subsampled} enhance subsampled Hessian-free Newton methods by combining the prior direction and the current Newton direction to form the search direction. \citet*{botev2017practical} introduce an efficient block-diagonal approximation to the Gauss-Newton matrix for training autoencoders. Finally, \citet*{wang2020newton} utilize Hessian-free Newton methods for training convolutional neural networks.

\section{Autoencoders} \label{autoencoders}
\begin{figure}[H]
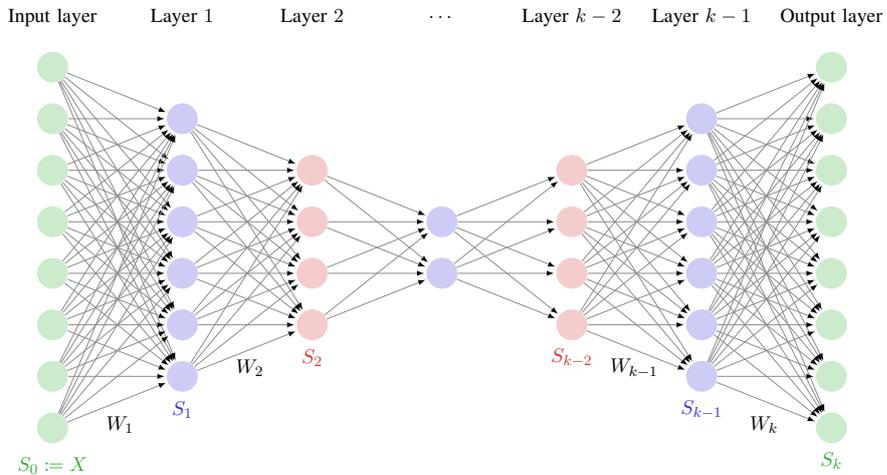

\centering
\begin{adjustbox}{width=0.75\textwidth}
\begin{neuralnetwork}[height=8]
  \tikzstyle{input neuron}=[neuron, fill=mygreen!20];
  \tikzstyle{hidden neuron}=[neuron, fill=myblue!20];
  \tikzstyle{output neuron}=[neuron, fill=myred!20];

  \inputlayer[count=8, bias=false, title=Input layer]

  \hiddenlayer[count=6, bias=false, title=Layer $1$]
  \linklayers
  
  \outputlayer[count=4, bias=false, title=Layer $2$]
  \linklayers

  \hiddenlayer[count=2, bias=false, title=$\cdots$]
  \linklayers

  \outputlayer[count=4, bias=false, title=Layer $k-2$]
  \linklayers
  
  \hiddenlayer[count=6, bias=false, title=Layer $k-1$]
  \linklayers

  \inputlayer[count=8, bias=false, title=Output layer]
  \linklayers
  \draw (1.3,-7.9) node{$W_1$};
  \draw (3.8,-6.8) node{$W_2$};
  \draw (11.2,-6.8) node{$W_{k-1}$};
  \draw (13.7,-7.9) node{$W_k$};
  \draw[mygreen!80] (0,-8.7) node{$S_0 \coloneqq X$};
  \draw[myblue!80] (2.5,-7.6) node{$S_1$};
  \draw[myred!80] (5,-6.6) node{$S_2$};
  \draw[myred!80] (10,-6.6) node{$S_{k-2}$};
  \draw[myblue!80] (12.5,-7.6) node{$S_{k-1}$};
  \draw[mygreen!80] (15,-8.6) node{$S_k$};

\end{neuralnetwork}
\end{adjustbox}
\caption{Deep autoencoder}
\label{fig:auto}
\end{figure}

Consider an autoencoder with $k$ non-input layers, as illustrated in Figure \ref{fig:auto}. Let $m_\ell$ denote the number of nodes in layer $\ell$. Given the set of training data $\left\{ \bm{x}_i \in \R^{m_1} \colon i = 1,\ldots,N \right\}$, we construct the input matrix for the training data:
\[
X_{\train} = 
\left[
\begin{array}{ccc}
\llongdash & \bm{x}_1^T & \rlongdash \\
\llongdash & \bm{x}_2^T & \rlongdash \\
 & \vdots & \\
\llongdash & \bm{x}_N^T & \rlongdash
\end{array}
\right] \in \R^{N \times m_1}.
\]
To minimize the computational cost of optimization, it is common practice to select a random subset of training points. Define $\mathcal{N} \coloneqq \{1,\ldots,N\}$, and let $\mathcal{S} \subseteq \mathcal{N}$ be a random sample consisting of $n$ indexes of training instances. The set $\left\{ \bm{x}_i \colon i \in \mathcal{S} \right\}$ of training points corresponding to the indexes in $\mathcal{S}$ is called a \textit{mini-batch}. The input matrix for the mini-batch is then constructed by selecting the rows corresponding to the indexes in $\mathcal{S}$:
\[
X \coloneqq X_{\train}[\mathcal{S},\, :\,] \in \R^{n \times m_1}.
\]
In the autoencoder, a bias term can be absorbed into a weight matrix by appending a vector of ones to the input matrix $X$. The augmented version of $X$ is  represented by
\[
\widetilde{X} \coloneqq 
\left[\begin{array}{cc} 
X & \bm{1} 
\end{array} \right] 
\in \R^{n \times (m_1+1)},
\]
where $\bm{1} \in \R^{n}$ is the vector of ones. In what follows, we adopt this notation (\,$\widetilde\cdot$\,) to represent matrices with a vector of ones of appropriate size appended. A weight matrix that connects layers $\ell$ and $\ell+1$ is denoted as $W_{\ell} \in \R^{(m_\ell+1) \times m_{\ell+1}}$. Combining all the weight matrices, we define $\bm{w} \coloneqq \big( W_1, \ldots,W_{k} \big)$. Given the layer structure $(m_1,\ldots,m_{k+1})$, the space of vectors of weight matrices can be defined as
\begin{equation} \label{W}
\mathcal{W} \coloneqq \Big\{ (W_1, W_2, \ldots, W_{k}) \colon W_\ell \in \R^{(m_\ell+1) \times m_{\ell+1}}, \ \ell=1,\ldots,k  \Big\}.
\end{equation}

\pagebreak
\noindent
Let $\widehat{\bm{x}}_i$ be the reconstructed vector of input vector $\bm{x}_i$. The reconstruction loss function for $\bm{x}_i$ is then defined by
\[
f_i(\bm{w}) \coloneqq f(\bm{w}; \bm{x}_i) \coloneqq \frac{1}{2}\bm{r}_i^T \bm{r}_i,
\]
where $\bm{r}_i \coloneqq \bm{r}(\bm{w}; \bm{x}_i) \coloneqq \widehat{\bm{x}}_i - \bm{x}_i$ represents the \textit{residual} vector for instance $i$. The goal is to minimize the mean squared errors for the training data, so we define the \textit{training loss function}:
\[
f_{\mathcal{N}}(\bm{w}) \coloneqq f(\bm{w}; X_{\train}) \coloneqq \frac{1}{N} \sum_{i=1}^N f(\bm{w}; \bm{x}_i) = \frac{1}{N} \sum_{i=1}^N f_i(\bm{w}).
\]
At each iteration of Hessian-free optimization, we choose a subset $\mathcal{S} \subseteq \mathcal{N}$ of size $n$ of the training dataset and apply one step of an optimization algorithm to the \textit{mini-batch loss function}:
\[
f_{\mathcal{S}}(\bm{w}) \coloneqq f(\bm{w}; X) \coloneqq \frac{1}{n} \sum_{i \in \mathcal{S}} f(\bm{w}; \bm{x}_i) = \frac{1}{n} \sum_{i \in \mathcal{S}} f_i(\bm{w}). 
\]
Consider the element-wise activation functions $\sigma_1, \ldots, \sigma_{k} \colon \R \to \R$ such that the weight matrix $W_\ell$ is followed by the activation function $\sigma_\ell$. For a single input vector $\bm{x}$, the $\ell^{\text{th}}$ non-input layer of the autoencoder can be expressed iteratively as
\begin{equation} \label{s_ell}
\bm{s}_{\ell} \coloneqq \sigma_\ell\!\left( W_\ell^T \widetilde{\bm{s}}_{\ell-1} \right), \qquad \ell = 1,\ldots,k,
\end{equation}
with $\bm{s}_0 \coloneqq \bm{x}$. The vector $\widetilde{\bm{s}}_\ell = (\bm{s}_\ell, 1)$ is obtained by appending $1$ to the vector $\bm{s}_\ell $. The output of non-input layer $\ell$ for instance $i$ is represented by $\bm{s}_{\ell,i}$, which can be written in terms of the weights $W_1,\ldots,W_{\ell}$ and the input vector $\bm{x}_i$ by unfolding the expression in \eqref{s_ell}. The reconstruction process of an input vector $\bm{x}_i$ is given below:
\[
\bm{x}_i^T \eqqcolon \bm{s}_{0,i}^T \ \ \xmapsto{W_1}\ \ \bm{s}_{1,i}^T \ \ \xmapsto{W_2}\ \ \bm{s}_{2,i}^T\ \ \xmapsto{W_3}\ \ \cdots\ \ \xmapsto{W_k}\ \ \bm{s}_{k,i}^T \eqqcolon \widehat{\bm{x}}_i^T.
\]
To create output matrices for layers $\ell=1,\ldots,k$, we combine the output vectors for mini-batch instances in $\mathcal{S}$. For notational brevity, we assume without loss of generality that $\mathcal{S} = \{1,\ldots,n\}$ and define the following output matrices, denoted by $S_\ell$, corresponding to $\mathcal{S}$:
\begin{align*}
\underbrace{\left[ \begin{array}{ccc}
\llongdash & \bm{x}_1^T & \rlongdash \\
\llongdash & \bm{x}_2^T & \rlongdash \\
 & \vdots & \\
\llongdash & \bm{x}_n^T & \rlongdash
\end{array} \right]}_{S_0 \coloneqq X} \xmapsto{W_1}
\underbrace{\left[ \begin{array}{ccc}
\llongdash & \bm{s}_{1,1}^T & \rlongdash \\
\llongdash & \bm{s}_{1,2}^T & \rlongdash \\
 & \vdots & \\
\llongdash & \bm{s}_{1,n}^T & \rlongdash
\end{array} \right]}_{S_1} \xmapsto{W_2}
\underbrace{\left[ \begin{array}{ccc}
\llongdash & \bm{s}_{2,1}^T & \rlongdash \\
\llongdash & \bm{s}_{2,2}^T & \rlongdash \\
 & \vdots & \\
\llongdash & \bm{s}_{2,n}^T & \rlongdash
\end{array} \right]}_{S_2} \xmapsto{W_3} \cdots \xmapsto{W_k}
\underbrace{\left[ \begin{array}{ccc}
\llongdash & \bm{s}_{k,1}^T & \rlongdash \\
\llongdash & \bm{s}_{k,2}^T & \rlongdash \\
 & \vdots & \\
\llongdash & \bm{s}_{k,n}^T & \rlongdash
\end{array} \right]}_{S_{k}}
\end{align*}
Letting $S_0 \coloneqq X$, the output matrix for non-input layer $\ell$ is given by
\begin{equation} \label{S}
S_\ell \coloneqq \sigma_\ell \big( \widetilde{S}_{\ell-1} W_\ell \big) \in \R^{n \times m_{\ell+1}}, \qquad \ell = 1,2,\ldots,k.
\end{equation}
As before, the vector of ones is appended to $S_\ell$ as the input matrix to the next layer. The augmented matrix is denoted by 
\[
\widetilde{S}_{\ell} \coloneqq \left[ \begin{array}{cc} S_\ell & \bm{1} \end{array} \right], 
\qquad \ell=0,\ldots,k-1.
\]
Next, given the input matrix $X$ for the mini-batch $\mathcal{S}$, we define the \textit{residual function} $R \colon \mathcal{W} \to \R^{n \times m_{1}}$ by
\[
R(\bm{w}; X) \coloneqq \frac{1}{\sqrt n}
\left[ \begin{array}{ccc}
\llongdash & \bm{r}_1^T & \rlongdash \\
\llongdash & \bm{r}_2^T & \rlongdash \\
 & \vdots & \\
\llongdash & \bm{r}_n^T & \rlongdash
\end{array} \right] =
\frac{1}{\sqrt{n}} \big[ S_{k} - S_0 \big]
\]
so that the mini-batch loss function $f(\bm{w}; X)$ can be written as
\[
f(\bm{w}; X) = \frac{1}{2} \big\| R(\bm{w}; X)\big\|_F^2.
\]

\subsection{Notation} \label{notation}
In deep neural networks, weight matrices are typically vectorized so that the parameters of interest can be written as a vector. Instead, we represent the weights as a vector of matrices, $\bm{w} = (W_1,\ldots, W_k)$, and create the space $\mathcal{W}$ of vectors of weight matrices in \eqref{W}. This approach allows us to utilize matrix algebra tools in Julia \citep{bezanson2012julia} to efficiently compute derivatives of $f$.

\vspace{2mm}\noindent
Hessian-free optimization involves computing the gradient of an objective function. To compute the gradient of $f$, we need the derivative of $R$ with respect to $\bm{w}$, which is denoted by $\nabla R(\bm{w}; X)$. The shape of $\nabla R(\bm{w}; X)$ is quite complex as $R$ maps a vector of matrices to a matrix. Note that the derivative of a matrix with respect to another matrix is a 4th-order tensor. Therefore, it is natural to express $\nabla R(\bm{w}; X)$ as a row vector of 4th-order-tensor-valued functions $\nabla_{W_\ell} R \colon \mathcal{W} \to \R^{n \times m_1 \times (m_\ell + 1) \times m_{\ell+1}}$; that is,
\[
\nabla R(\bm{w}; X) =
\left[
\begin{array}{cccc}
\nabla_1 R & \nabla_2 R & \cdots & \nabla_{k} R
\end{array}
\right],
\]
where $\nabla_\ell R$ stands for $\nabla_{W_\ell} R(\bm{w}; X)$. Now, let us recall the double dot products for tensors. The double dot product of $4$th-order tensor $A$ and $2$th-order tensor (i.e. matrix) $B$ is defined by
\[
(A \dd B)_{pq} \coloneqq \sum\nolimits_{r,s} A_{pqrs} B_{rs}.
\]
The double dot product of $4$th-order tensors $A$ and $B$ is defined by
\[
(A \dd B)_{pqrs} \coloneqq \sum\nolimits_{i,j} A_{pqij} B_{ijrs}.
\]
The adjoint of a $4$th-order tensor $A$, denoted by $A^T$, is defined by $(A^T)_{pqrs} = A_{rspq}$.

\vspace{2mm}\noindent
For matrices $A$ and $B$ of the same size, $A \circ B$ denotes the \textit{Hadamard} (or \textit{entrywise}) \textit{product} of $A$ and $B$; that is, $(A \circ B)_{ij} \coloneqq A_{ij}B_{ij}$; and $A \bullet B$ denotes the \textit{Frobenius inner product} of $A$ and $B$; that is,
\[
A \bullet B \coloneqq \sum\nolimits_{i,j} A_{ij}B_{ij} = \trace(A^TB).
\]
The \textit{Frobenius norm} of a matrix $A$ (not necessarily square) is defined by $\|A\|_F \coloneqq \sqrt{A \bullet A}$. For notational brevity, we assume that the Hadamard product operator $\circ$ has higher precedence than the Frobenius inner product $\bullet$ and lower precedence than ordinary matrix multiplication. Thus, 
\begin{itemize}
\item[--] $A \bullet B \circ C$ is interpreted as $A \bullet (B \circ C)$, and
\item[--] $A \circ BC$ is interpreted as $A \circ (BC)$.
\end{itemize}
To work with $\nabla R(\bm{w}; X)$, we need a space of matrices of 4th-order tensors. Given positive integers $k, \ell$ and vectors $p, q \in \N^k$ and $r, s \in \N^{\ell}$, the space of matrices of 4th-order tensors is defined as
\[
\mathcal{A}_{k,\ell}^{p,q,r,s} \coloneqq
\left\{
\left[
\begin{array}{cccc}
A_{11} & A_{12} & \cdots & A_{1\ell} \\
A_{21} & A_{22} & \cdots & A_{2\ell} \\
\vdots & \vdots & \ddots & \vdots \\
A_{k1} & A_{k2} & \cdots & A_{k\ell} 
\end{array}
\right] \colon A_{ij} \in \R^{p_{i} \times q_{i} \times r_{j} \times s_{j}}
\right\}.
\]
For $A \in \mathcal{A}_{k,\ell}^{p,q,r,s}$ and $B \in \mathcal{A}_{\ell,m}^{r,s,t,u}$, we define the product $AB \in \mathcal{A}_{k,m}^{p,q,t,u}$ by
\[
(AB)_{ij} = \sum_{h=1}^\ell A_{ih} \dd B_{hj} \in \R^{p_i \times q_i \times t_j \times u_j}, \qquad i=1,\ldots,k; \quad j=1,\ldots,m.
\]
The adjoint of $A \in \mathcal{A}_{k,\ell}^{p,q,r,s}$ is defined as
\[
A^T \coloneqq 
\left[
\begin{array}{cccc}
A_{11}^T & A_{21}^T & \cdots & A_{k1}^T \\
A_{12}^T & A_{22}^T & \cdots & A_{k2}^T \\
\vdots & \vdots & \ddots & \vdots \\
A_{1\ell}^T & A_{2\ell}^T & \cdots & A_{k\ell}^T
\end{array}
\right] \in \mathcal{A}_{\ell,k}^{r,s,p,q}.
\]
Now, using the notation above, we write four spaces of interest. Recall that $k$ is the number of non-input layers in the network, $n$ is the mini-batch size, and $m_1,\ldots,m_{k+1}$ are the number of nodes in the layers. Define $r \coloneqq (m_1+1, m_2+1, \ldots, m_k+1)$ and $s \coloneqq (m_2, m_3, \ldots, m_{k+1})$. Given $k, n, m_1, \ldots, m_{k+1}$, 
\begin{itemize}
\item[(i)] the space of input matrices $X$ can be written as $\mathcal{X} \coloneqq \mathcal{A}_{1,1}^{n,m_1,1,1} = \R^{n \times m_1}$,
\item[(ii)] the space of weights $\bm{w}$ can be written as $\mathcal{W} \coloneqq \mathcal{A}_{k,1}^{r,s,1,1}$,
\item[(iii)] the space of derivatives of $R$ with respect to $\bm{w}$ can be written as $\mathcal{J} \coloneqq \mathcal{A}_{1,k}^{n,m_1,r,s}$, and
\item[(iv)] the space of Gauss-Newton matrices $G$ can be written as $\mathcal{G} \coloneqq \mathcal{A}_{k,k}^{r,s,r,s}$.
\end{itemize}
Since the space $\W$ of weights is of special interest, we define two operations on $\W$: for $\bm{v}, \bm{w} \in \W$ and $\alpha \in \R$,
\[
\bm{v} + \bm{w} = 
\left[ \begin{array}{c}
V_1 \\ V_2 \\ \vdots \\ V_k
\end{array} \right] +
\left[ \begin{array}{c}
W_1 \\ W_2 \\ \vdots \\ W_k
\end{array} \right] \coloneqq 
\left[ \begin{array}{c}
V_1 + W_1 \\ V_2 + W_2 \\ \vdots \\ V_k + W_k
\end{array} \right], \qquad 
\alpha\, \bm{v} \coloneqq 
\left[ \begin{array}{c}
\alpha V_1 \\ \alpha V_2 \\ \vdots \\ \alpha V_k
\end{array} \right].
\]
Notice that $\W$ is a vector space over $\R$. Now, we define the inner product $\langle \cdot, \cdot \rangle \colon \W \times \W \to \R$ by
\[
\langle \bm{a}, \bm{b} \rangle \coloneqq \bm{a}^T \bm{b} = \sum_{\ell=1}^{k} (A_\ell \bullet B_\ell).
\]
Consider the norm $\|\cdot\|_{2}$ induced by the inner product $\langle \cdot, \cdot \rangle$ as follows:
\[
\|\bm{w}\|_{2} \coloneqq \sqrt{\langle \bm{w}, \bm{w} \rangle} = \left[ \sum_{\ell=1}^k \left\|W_\ell \right\|_F^2 \right]^{1/2}, \qquad \forall \bm{w} \in \mathcal{W}.
\]
We further define the norm $\| \cdot \|_1$ for the space $\W$ as follows:
\[
\|\bm{w}\|_1 \coloneqq \sum_{\ell=1}^k \|W_\ell\|_{1,1} = \sum_{\ell=1}^k \sum_{i,j} |(W_\ell)_{ij}|, \qquad \forall \bm{w} \in \W.
\]
Finally, we define the Hadamard product of vectors $\bm{v}, \bm{w} \in \mathcal{W}$ by
\[
\bm{v} \circ \bm{w} =
\left[ \begin{array}{c}
V_1 \\ V_2 \\ \vdots \\ V_k
\end{array} \right] \circ
\left[ \begin{array}{c}
W_1 \\ W_2 \\ \vdots \\ W_k
\end{array} \right] \coloneqq 
\left[ \begin{array}{c}
V_1 \circ W_1 \\
V_2 \circ W_2 \\
\vdots \\
V_k \circ W_k
\end{array} \right].
\]

\section{Stochastic Hessian-free optimization} \label{SHF}
This section discusses Hessian-free optimization \citep{martens2010deep} and presents our contribution to HF optimization in a stochastic setting. Consider a nonlinear least squares problem:
\[
f(\bm{w}; X) = \frac{1}{2} \big\| R(\bm{w}; X) \big\|_F^2,
\]
where the residual function $R$ is assumed to be twice continuously differentiable. We model the objective function $f$ using the local quadratic approximation around $\bm{w}$:
\begin{equation} \label{approx}
f(\bm{w} + \bm{d}; X) \approx f(\bm{w}; X) + \bm{d}^T \nabla f(\bm{w}; X) + \frac{1}{2} \bm{d}^T H \bm{d},
\end{equation}
where $\bm{d} \in \mathcal{W}$ is a search direction and $H \in \mathcal{G}$ is a curvature matrix. In Newton's method, the matrix $H$ is chosen to be the Hessian of $f$ at $\bm{w}$. The Gauss-Newton matrix is obtained by approximating the Hessian of $f$ by eliminating the term involving second derivatives of $R(\bm{w}; X)$; that is,
\[
\Hess f(\bm{w}; X) \approx \nabla R(\bm{w}; X)^T \nabla R(\bm{w}; X).
\]
Since we do not entirely trust the quadratic model in \eqref{approx} as an approximation for large values of $\bm{d}$, we add a damping term $\lambda^2 I$ to the approximation of the Hessian for some constant $\lambda \geq 0$:
\begin{equation} \label{Hess}
\Hess f(\bm{w}; X) \approx \nabla R(\bm{w}; X)^T \nabla R(\bm{w}; X) + \lambda^2 I,
\end{equation}
which is called the \textit{regularized} Gauss-Newton matrix. Note that the Hessian approximation in \eqref{Hess} guarantees the positive definiteness, which ensures the existence of a minimizer. The gradient of $f$ at $\bm{w}$ is given by 
\begin{equation} \label{grad}
\nabla f(\bm{w}; X) = \nabla R(\bm{w}; X)^T R(\bm{w}; X).
\end{equation}
Thus, by plugging \eqref{Hess}\,-\,\eqref{grad} into \eqref{approx} and dropping the constant term $f(\bm{w}; X)$, the regularized Gauss-Newton method solves
\begin{equation} \label{GN}
\min_{\bm{d} \in \mathcal{W}}\ \left[ \bm{d}^T\nabla R(\bm{w}; X)^T R(\bm{w}; X) + \frac{1}{2} \bm{d}^T \nabla R(\bm{w}; X)^T \nabla R(\bm{w}; X)\,\bm{d} + \frac{\lambda^2}{2}\!\left\|\bm{d} \right\|_2^2 \right].
\end{equation}
Here, the damping term $\frac{\lambda^2}{2}\!\left\|\bm{d} \right\|_2^2$ can be interpreted as an $\ell_2$-penalty, which helps us avoid overfitting. The regularized Gauss-Newton direction is the minimizer $\bm{d}^*$ of \eqref{GN}, which can be found by solving the normal equations:
\begin{equation} \label{normal}
\Big( \nabla R(\bm{w}; X)^T \nabla R(\bm{w}; X) + \lambda^2 I \Big) \bm{d}^* = -\nabla R(\bm{w}; X)^T R(\bm{w}; X).
\end{equation}
Once the search direction $\bm{d}^*$ is determined, the step-length $s$ is chosen via Armijo/backtracking line search; that is to find $s \in (0,1]$ that satisfies the sufficient decrease condition:
\[
f(\bm{w} + s\bm{d}^*; X) \leq f(\bm{w}; X) + \alpha\,s\,\nabla f(\bm{w}; X)^T \bm{d}^*,
\]
where $\alpha \in (0,1)$ is chosen fairly small. The weight vector is then updated as $\bm{w} \gets \bm{w} + s\bm{d}^*$.

\vspace{2mm}\noindent
It is worth noting that HF optimization does not require second derivatives. We only need to compute the gradient $\nabla R(\bm{w}; X)$. The system \eqref{normal} involves a large number of variables, denoted by $n_{\text{var}} \coloneqq \sum_{\ell=1}^{k} (m_\ell +1)m_{\ell+1}$. This number can be huge, making it impossible to store the $n_{\text{var}} \times n_{\text{var}}$ Hessian matrix, let alone invert it. Therefore, the cost of solving the system \eqref{normal} can be very high. To overcome this, we use the LSMR method, which is explained in detail in Section \ref{LSMR}.

\vspace{2mm}\noindent
Our implementation of stochastic HF optimization is outlined as pseudo-code in Algorithm \ref{main}. Section \ref{weight_init} introduces a weight initialization that demonstrates faster convergence in our design. In Section \ref{LM_heuristic}, the Levenberg-Marquardt heuristic is provided for dynamically adjusting the damping parameter $\lambda$ to adapt to changes in the local curvature of $f$. In Section \ref{mini-batch_size}, we explore reducing the data used in training to accelerate the optimization process. Section \ref{precon} describes how to reduce the number of LSMR iterations using preconditioning. The approaches for initializing LSMR and determining the number of iterations in LSMR are discussed in Section \ref{init_LSMR}. Finally, Section \ref{matrix-vector_prods} explains an efficient procedure for computing the matrix-vector products.

\begin{table}[H]
{\LinesNumberedHidden
\begin{algorithm}[H] \label{main} 
\caption{Stochastic Hessian-Free Optimization}
\SetKwProg{fnc}{function}{}{end}
\SetKwInOut{Given}{Given}
\SetAlgorithmName{Algorithm}{}

\Given{$X_{\train}$;\ \ $X_{\val}$;\ \ $m_0, n_1, \maxiter_1, n_{\max} \in \N$;\ \ $\mathrm{drop}, \alpha, \gamma_1, \theta \in (0,1)$;\ \ $
\sigma, \lambda_1 \geq 0$; $\atol, \ftol$}
Initialize $\bm{w}$ according to \eqref{Winit} with $m_0, \sigma$; \quad $\bm{d} \gets \bm{0}$\tcp*{Section \ref{weight_init}}
$n \gets n_1$; \quad $\lambda \gets \lambda_1$; \quad $\gamma \gets \gamma_1$; \quad $\maxiter \gets \maxiter_1$; \quad $i \gets 1$\;
$N \gets $ number of instances (rows) in $X_{\train}$\;
\While{$\mathrm{not\ converged}$}
{
$\mathcal{S} \gets$ draw $n$ instances from $\{1,\ldots,N\}$ randomly; \quad
$X \gets X_{\train}[\mathcal{S}, :]$\;
$A \gets \nabla R(\bm{w}; X)$;\quad $\bm{b} \gets -R(\bm{w}; X)$; \quad $\bm{c} \gets \mathrm{Precon}(\bm{w}; X)$\tcp*{Section \ref{precon}}
Define the merit function $\phi(\bm{d}) \coloneqq f(\bm{w}+\bm{d}; X_{\val})$\;
Solve $(A^TA + \lambda^2 I)\bm{d} = A^T\bm{b}$ for $\bm{d}$ using $\LSMR(\bm{d}, A, \bm{b}, \lambda, \bm{c}; \maxiter, \atol, \ftol, \phi)$\;
$\rho \gets \dfrac{f(\bm{w} + \bm{d}; X) - f(\bm{w}; X)}{\frac{1}{2}\bm{d}^TA^TA\bm{d} - \bm{d}^T A^T\bm{b}}$\tcp*{Section \ref{LM_heuristic}}
\textbf{if} $\rho < 0.25$ \textbf{then} $\lambda \gets \lambda/\mathrm{drop}$; \textbf{elseif} $\rho > 0.75$ \textbf{then} $\lambda \gets \mathrm{drop} \cdot \lambda$ \textbf{end}\\
\tcp{Backtracking line search}
$s \gets 1$\;
\While{$f(\bm{w} + s\bm{d}; X) > f(\bm{w}; X) - \alpha\,s\,\bm{d}^T A^T \bm{b}$}
{
	$s \gets s/2$\;
}
$\bm{w} \gets \bm{w} + s\bm{d}$\;
$\bm{d} \gets \gamma \bm{d}$; \quad $\gamma \gets \min(1.002 \cdot \gamma,\ 0.95)$\tcp*{Section \ref{init_LSMR}}
$f_{\val,i} \gets f(\bm{w}; X_{\val})$\;
$\widehat{n}_i \gets \mathrm{BatchSize}(\bm{w}, \mathcal{S}, N, \theta)$\tcp*{Section \ref{mini-batch_size}}
\If{$i > 5$}
{
$r_{\rel} \gets \frac{f_{\val,i-5} - f_{\val,i}}{f_{\val,i}}$; \quad
$\widehat{n}_{\avg} \gets \left\lceil \frac{\widehat{n}_{i-4} + \widehat{n}_{i-3} + \cdots + \widehat{n}_{i}}{5} \right\rceil$\;
\uIf{$\widehat{n}_{\avg} > n$}{
$n_{\mathrm{next}} \gets \min(\widehat{n}_{\avg}, n_{\max})$\;
}
\uElseIf{$r_{\rel} < 0.005$}{
$n_{\mathrm{next}} \gets \min(\lceil 1.005 \cdot n \rceil, n_{\max})$\;
}
\Else{$n_{\mathrm{next}} \gets n$\;}
$\mathrm{maxiter} \gets \left\lceil\frac{n_{\mathrm{next}}}{n} \cdot \mathrm{maxiter} \right\rceil$; \quad $n \gets n_{\mathrm{next}}$\tcp*{Section \ref{init_LSMR}}
}
$i \gets i+1$\;
}
\end{algorithm}}
\end{table}

\subsection{Weight initialization} \label{weight_init}
From all the weight initialization methods we have tried, we found that \citeauthor{martens2010deep}' sparse initialization scheme yields the best performance. To implement this scheme, we restrict the number of nonzero incoming connection weights to each unit to a specific number, denoted by $m_0$, and we set the biases to $0$. That is,
\begin{equation} \label{Winit}
\left.\begin{cases}
(W_\ell)_{ij} \sim N(0, \sigma) & \quad \text{if } i \in \mathcal{I}_{j\ell} \\
(W_\ell)_{ij} = 0 & \quad \text{if } i \notin \mathcal{I}_{j\ell}
\end{cases} \right\},
\qquad j = 1,\ldots,m_{\ell+1}, \quad \ell =1,\ldots,k,
\end{equation}
where $\mathcal{I}_{j\ell}$ is the set of indexes of size $\min(m_0, m_\ell)$ randomly selected from the set $\{1,\ldots,m_\ell\}$. In our experiments, we set $m_0 = 10$ for the MNIST and USPS datasets and $m_0=20$ for the CURVES dataset, and $\sigma=1.5$.

\subsection{Levenberg-Marquardt heuristic} \label{LM_heuristic}
The regularization (or damping) parameter $\lambda$ is adjusted at each iteration based on the agreement between the objective function $f$ and its quadratic approximation. Specifically, we define the ratio between the actual reduction in $f$ and the predicted reduction obtained by the quadratic model in \eqref{approx} as follows:
\[
\rho \coloneqq \frac{f(\bm{w} + \bm{d};X) - f(\bm{w};X)}{\bm{d}^T\nabla f(\bm{w};X) + \frac{1}{2} \bm{d}^T\nabla R(\bm{w};X)^T\nabla R(\bm{w}; X) \bm{d}}.
\]
The parameter $\lambda$ is useful for adjusting the search directions between gradient descent and Gauss-Newton directions in an adaptive way. When $\lambda$ has a smaller value, the step taken is closer to the Gauss-Newton direction, while larger values of $\lambda$ result in a step closer to the negative gradient.

\vspace{2mm}\noindent
The Levenberg-Marquardt heuristic has been widely used for HF optimization. It updates the parameter $\lambda$ according to the following rule:
\[
\lambda \gets 
\begin{cases}
\lambda/\mathrm{drop} & \text{if } \rho < \frac{1}{4}, \\[0.2em]
\mathrm{drop} \cdot \lambda & \text{if } \rho > \frac{3}{4},
\end{cases}
\]
where $\mathrm{drop} \in (0, 1)$ is a user-defined constant to control the drop in $\lambda$. When $\rho$ is relatively small, i.e., $\rho < 1/4$, the quadratic model overestimates the reduction in $f$. Thus, the value of $\lambda$ is enlarged for the next iteration, resulting in smaller future updates that are more likely to be contained within the region where the quadratic model accurately estimates the reduction. On the other hand, when $\rho$ is close to $1$, i.e., $\rho > 3/4$, the quadratic approximation of $f$ is expected to be quite accurate near $\bm{d}^*$. Therefore, it is affordable to reduce the value of $\lambda$, which produces larger and more aggressive updates towards the Newton direction. 

\vspace{2mm}\noindent
Regarding the values of $\mathrm{drop}$, \citet{martens2010deep} uses $\mathrm{drop} = 2/3$, whereas \citet{kiros2013training} prefers a softer damping criterion with $\mathrm{drop} = 99/100$. \citet{kiros2013training} observes that \citeauthor{martens2010deep}' damping criterion is too harsh in the stochastic setting as $\lambda$ frequently oscillates. Our experimental results align with \citeauthor{kiros2013training}' observation. We find that $\mathrm{drop} = 49/50$ or $99/100$ consistently leads to a smooth decrease in $\lambda$ during the optimization process.

\vspace{2mm}\noindent
When it comes to initializing the parameter $\lambda$, the optimal value can vary depending on the problem at hand. However, it is generally a good idea to start with a larger value for $\lambda$ so that the initial updates are small and move in the direction of the gradient descent. As the training progresses, $\lambda$ should converge to zero with a suitable initial value. In our experiments, we typically initialize the parameter $\lambda$ with a value of 10.

\subsection{Mini-batch size selection} \label{mini-batch_size}
The main objective of this paper is to reduce the amount of data used for training in order to accelerate the optimization process. Our algorithm begins with a small subset of the training data, known as a mini-batch, and then progressively increases the mini-batch size based on (i) variance estimates acquired during the computation of a mini-batch gradient and (ii) a relative decrease in the objective value for the validation data.

\vspace{2mm}\noindent
The algorithm proposed by \citet*{byrd2012sample} suggests increasing the mini-batch size based on variance estimates obtained during the computation of a mini-batch gradient. At each HF iteration, a subset $\mathcal{S}$ of $\mathcal{N}$ is chosen, and one step of an optimization algorithm is applied to the batch objective function $f_{\mathcal{S}}(\bm{w})$. The algorithm starts with a relatively small mini-batch size and increases the size throughout the progression of the algorithm. Using a small initial mini-batch leads to rapid progress in the early stages, while a larger mini-batch becomes helpful in providing high accuracy in the solution during later training iterations.

\vspace{2mm}\noindent
The algorithm not only enforces descent in $f_{\mathcal{S}}$ at every iteration but also ensures a descent direction for the training objective function $f_{\mathcal{N}}$. The gradients of the mini-batch and training losses are represented by $\nabla f_{\mathcal{S}}(\bm{w})$ and $\nabla f_{\mathcal{N}}(\bm{w})$, respectively:
\[
\nabla f_{\mathcal{S}}(\bm{w}) = \frac{1}{n} \sum_{i \in \mathcal{S}} \nabla f_i(\bm{w}), \qquad 
\nabla f_{\mathcal{N}}(\bm{w}) = \frac{1}{N} \sum_{i \in \mathcal{N}} \nabla f_i(\bm{w}),
\]
where $n = |\mathcal{S}|$ and $N = |\mathcal{N}|$. 
The vector $\bm{d} = -\nabla f_{\mathcal{S}}(\bm{w})$ is a descent direction for $f_{\mathcal{N}}$ at $\bm{w}$ if 
\begin{equation} \label{deltadef}
\delta_{\mathcal{S}}(\bm{w}) \coloneqq \left\|\nabla f_{\mathcal{S}}(\bm{w}) - \nabla f_{\mathcal N}(\bm{w})\right\|_2 \leq \theta \left\|\nabla f_{\mathcal{S}}(\bm{w}) \right\|_2,
\end{equation}
where $\theta \in (0,1)$. However, the quantity $\delta_{\mathcal{S}}(\bm{w})$ is not available as the evaluation of $\nabla f_{\mathcal{N}}(\bm{w})$ is computationally expensive. Thus, \citet{byrd2012sample} proposes approximating $\delta_{\mathcal{S}}(\bm{w})$ by an estimate of the variance of the random vector $\nabla f_{\mathcal{S}}(\bm{w})$. For a given $\bm{w} \in \mathcal{W}$, the expected value $\nabla f_{\mathcal{S}}(\bm{w})$ over all possible samples $\mathcal{S} \subseteq \mathcal{N}$ is  $\nabla f_{\mathcal{N}}(\bm{w})$. This implies that
\begin{equation} \label{delta}
\E\!\left[ \delta_{\mathcal{S}}(\bm{w})^2 \right] = \E\!\left[ \big\| \nabla f_{\mathcal{S}}(\bm{w}) - \nabla f_{\mathcal N}(\bm{w}) \big\|_2^2 \right]
= \left\| \Var\!\left(\nabla f_{\mathcal{S}}(\bm{w}) \right) \right\|_1,
\end{equation}
where the variance and square are taken component-wise. Note that $\Var(\nabla f_{\mathcal{S}}(\bm{w})) \in \W$. The variance of $\nabla f_{\mathcal{S}}(\bm{w})$ with a sample $\mathcal{S}$ chosen without replacement is given by 
\begin{equation} \label{vardf}
\Var\!\left( \nabla f_{\mathcal{S}}(\bm{w}) \right) = \frac{\Var(\nabla f_i(\bm{w}))}{n} \cdot \frac{N-n}{N-1}.
\end{equation}
It follows from \eqref{delta} and \eqref{vardf} that
\[
\E\!\left[ \delta_{\mathcal{S}}(\bm{w})^2 \right] \approx \frac{\left\| \Var_{i \in \mathcal{S}}\!\left(\nabla f_i(\bm{w}) \right) \right\|_1}{n} \cdot \frac{N-n}{N-1}.
\]
The population variance $\Var(\nabla f_i(\bm{w}))$ can be approximated with the sample variance
\[
\Var_{i \in \mathcal{S}}\left( \nabla f_i(\bm{w}) \right) = \frac{n}{n-1} \left[ \frac{1}{n}\sum_{i \in \mathcal{S}} (\nabla f_i(\bm{w}))^2 - (\nabla f_{\mathcal{S}}(\bm{w}))^2 \right],
\]
where the squares are again taken component-wise. Hence, the condition \eqref{deltadef} reduces to 
\begin{equation} \label{ineq}
\frac{\left\| \Var_{i \in \mathcal{S}}\!\left(\nabla f_i(\bm{w}) \right) \right\|_1}{n} \cdot \frac{N-n}{N-1} \leq \theta^2 \left\|\nabla f_{\mathcal{S}}(\bm{w}) \right\|_2.
\end{equation}
If inequality \eqref{ineq} is not satisfied, we increase the sample size $n$ to $\widehat{n}$ so that the descent condition is satisfied. Here is the heuristics used by \citet{byrd2012sample}. We want to find a larger sample $\widehat{S}$. Assume that the change in sample size is small enough that for any $\bm{w} \in \W$ we have
\[
\left\| \Var_{i \in \widehat{\mathcal{S}}}\!\left(\nabla f_i(\bm{w}) \right) \right\|_1 \approx
\left\| \Var_{i \in \mathcal{S}}\!\left(\nabla f_i(\bm{w}) \right) \right\|_1 \quad \text{and} \quad
\left\|\nabla f_{\widehat{\mathcal{S}}}(\bm{w}) \right\|_2 \approx
\left\|\nabla f_{\mathcal{S}}(\bm{w}) \right\|_2.
\]
Under these assumptions, condition \eqref{ineq} is satisfied if a new sample $\widehat{S}$ with size $\widehat{n}$ is chosen so that
\begin{equation} \label{descent.cond}
\frac{\left\| \Var_{i \in \mathcal{S}}\!\left(\nabla f_i(\bm{w}) \right) \right\|_1}{\widehat{n}} \cdot \frac{N-\widehat{n}}{N-1} \leq \theta^2 \left\|\nabla f_{\mathcal{S}}(\bm{w}) \right\|_2.
\end{equation}
Hence, we select a new mini-batch whose size is given by
\[
\widehat{n} = \left\lceil \frac{N \left\| \Var_{i \in \mathcal{S}}\!\left(\nabla f_i(\bm{w}) \right) \right\|_1}{\left\| \Var_{i \in \mathcal{S}}\!\left(\nabla f_i(\bm{w}) \right) \right\|_1 + \theta^2 (N-1) \left\|\nabla f_{\mathcal{S}}(\bm{w}) \right\|_2} \right\rceil.
\]
Note that we do not assume $N \gg n$ unlike \citet{byrd2012sample}, which is why our formula is more complex than theirs. The algorithm for selecting mini-batch size is described in Algorithm \ref{algo:batch}.

\begin{table}[H]
{\LinesNumberedHidden
\begin{algorithm}[H] \label{algo:batch} 
\caption{Mini-batch size}
\SetKwProg{fnc}{function}{}{end}
\SetKwInOut{Given}{Given}
\SetAlgorithmName{Algorithm}{}

\fnc{$\mathrm{BatchSize}(\bm{w}, \mathcal{S}, N, \theta)$}
{
$n \gets |\mathcal{S}|$\;
$\nabla f_{\mathcal{S}}(\bm{w}) \gets \frac{1}{n} \sum_{i \in \mathcal{S}} \nabla f_{i}(\bm{w})$\;
$\Var(\nabla f_{\mathcal{S}}(\bm{w})) \gets \frac{n}{n-1} \left[ \frac{1}{n}\sum_{i \in \mathcal{S}} (\nabla f_i(\bm{w}))^2 - (\nabla f_{\mathcal{S}}(\bm{w}))^2 \right]$\;
$\widehat{n} \gets \left\lceil \dfrac{N \left\| \Var(\nabla f_{\mathcal{S}}(\bm{w})) \right\|_1}{\left\| \Var(\nabla f_{\mathcal{S}}(\bm{w})) \right\|_1 + \theta^2 (N-1) \left\|\nabla f_{\mathcal{S}}(\bm{w}) \right\|_2} \right\rceil$\;
\Return $\widehat{n}$
}
\end{algorithm}}
\end{table}

\noindent
After predicting the mini-batch size $\widehat{n}_i$ for iteration $i$ that satisfies the descent condition \eqref{descent.cond}, we do not implement it immediately in the next iteration. Instead, we take the average of the last five mini-batch size predictions to prevent drastic changes and smooth out the mini-batch size.
\[
\widehat{n}_{\avg} = \left\lceil \frac{\widehat{n}_{i-4} + \widehat{n}_{i-3} + \cdots + \widehat{n}_i}{5} \right\rceil.
\]
We observe that the relative decrease in the objective value becomes smaller in later HF iterations. In order to improve the convergence rate, we also adjust the mini-batch size based on the relative decrease in the objective value for the validation dataset. We take into account the relative decrease over the last five iterations:
\[
r_{\rel,i} = \frac{f_{\val,i-4} - f_{\val,i}}{f_{\val,i}}.
\]
If $r_{\rel,i} < 0.005$, the mini-batch size is increased by the factor of $\eta > 1$. To prevent mini-batch sizes from blowing up, we set the maximum sample size for mini-batches, denoted by $n_{\max}$. 

\vspace{2mm}\noindent
The parameter $\theta$ determines how much variation in the mini-batch gradient $\nabla f_{\mathcal{S}}$ the algorithm can tolerate. A higher value of $\theta$ means the algorithm is more tolerant of larger variances and is likely to keep a mini-batch size for longer. In addition, \citet{byrd2012sample} notes that larger values of $\theta$ can improve overall performance by allowing smaller mini-batch sizes in the initial iterations of the algorithm. Our experimental results confirm this observation. On the other hand, it is worth noting that as $\theta$ increases, the behavior of the algorithm may become more erratic.

\subsection{Preconditioner} \label{precon}
Preconditioning is a technique used in linear systems to make them easier to solve. In HF optimization, preconditioning is widely used to reduce the number of iterations required to estimate the Hessian implicitly. However, finding an appropriate preconditioner can be challenging for two reasons. First, the type of preconditioner that works best depends on the problem at hand. Second, obtaining a suitable preconditioner can be computationally expensive, which means that the cost of obtaining a good one may outweigh its benefits. Due to these limitations, any preconditioner that is more complex than Jacobi preconditioners is computationally infeasible for our problem.

\vspace{2mm}\noindent
In our research, we employ the Jacobi preconditioner proposed by \citet*{chapelle2011improved}, specifically designed for HF methods in deep autoencoders. This preconditioner is a randomized algorithm that provides an unbiased estimate of the diagonal of the generalized Gauss-Newton matrix. It requires the same level of effort as computing the gradient. Additionally, we test the diagonal preconditioner created by \citet{martens2010deep}, which utilizes the empirical Fisher information matrix. \citet{chapelle2011improved} argue that their approach optimizes the reconstruction error of a deep autoencoder's objective function more efficiently and effectively, which aligns with the results of our experiments.

\vspace{2mm}\noindent
The LSMR method solves the problem
\[
\min_{\bm{x}}\, \left\| A\,\bm{x} -\bm{b} \right\|_2^2 + \lambda^2 \left\|\bm{x}\right\|_2^2,
\]
which has the normal equation $(A^TA + \lambda^2 I) \bm{x} = A^T\bm{b}$. Using a preconditioner $C$, the LSMR method solves the transformed problem
\[
\min_{\bm{y}}\, \left\| AC^{-1}\bm{y} -\bm{b} \right\|_2^2 + \lambda^2 \|\bm{x}\|_2^2,
\]
whose normal equation is given by $\left( C^{-T}A^TAC^{-1} + \lambda^2 I \right)\bm{y} = C^{-T}A^T\bm{b}$, and then recovers the step size $\bm{x}$ by solving $\bm{y} = C\bm{x}$. Now, we need to code matrix-vector products of the form $\bm{v} \mapsto AC^{-1}\bm{v}$ and $\bm{u} \mapsto C^{-T}A^T \bm{u}$ by solving linear systems involving $C$ and $C^T$ before and after the products with $A$ and $A^T$, respectively.

\vspace{2mm}\noindent
With the notation introduced previously, we have $A = \nabla R(\bm{w}; X)$, $\bm{b} = -R(\bm{w}; X)$ and $\bm{x} = \bm{d} \in \mathcal{W}$. The preconditioner $C \in \mathcal{G}$ is diagonal; that is, only the diagonal entries $(C_{\ell\ell})_{pq,pq}$ are nonzero. Hence, we define $\bm{c} \coloneqq (C_1,\ldots,C_k) \in \mathcal{W}$ to store the reciprocals of the diagonal entries of $C$; that is, $(C_\ell)_{pq} = 1/(C_{\ell\ell})_{pq,pq}$. Therefore, we may write the matrix-vector products as $\bm{d} \mapsto A(\bm{c} \circ \bm{d})$ and $U \mapsto \bm{c} \circ A^TU$.

\vspace{2mm}\noindent
Recall that the Gauss-Newton matrix in our problem is
\[
G \coloneqq A^TA = \nabla R(\bm{w}; X)^T \nabla R(\bm{w}; X) \in \mathcal{G}.
\]
We want to estimate the diagonal elements of $G$. Let $(\ell, p, q)$ be indexes to refer to as the weight $(W_\ell)_{pq}$. So, a typical diagonal element of $G$ is given by
\[
(G_{\ell\ell})_{pq,pq} = \sum_{i=1}^n \sum_{j=1}^{m_1} \big( \nabla_\ell R \big)_{ij,pq}^2.
\]
Here, index $\ell$ stands for the layer, index $i$ for the training instance, index $j$ for the input/output unit. Next, we define the following random variable:
\[
c_{\ell pq,i} = \sum_{j=1}^{m_1} \big( \nabla_\ell R \big)_{ij,pq} u_{ij},
\]
where $u_{ij}$ is a random variable that takes on values $\pm 1$ with equal probability. \citet*{chapelle2011improved} prove that $\sum_{i=1}^n c_{\ell pq,i}^2$ is an unbiased estimate of $(G_{\ell\ell})_{pq,pq}$. Moreover, if the training data is i.i.d., its relative variance decays as $\mathcal{O}(1/n)$, where $n$ is the number of mini-batch instances. In order to rule out the possibility of zero denominator and for stability, it is beneficial to include an additional additive constant $1$ to the estimate $\sum_{i=1}^n c_{\ell pq,i}^2$. Hence, we fill out the entries of the inverse of the preconditioner as follows:
\[
(C_\ell)_{pq} \coloneqq \frac{1}{1+ \sum_{i=1}^n c_{\ell pq,i}^2}.
\]

\vspace{2mm}\noindent
The inverse $\bm{c}$ of the preconditioner can be computed using backward passes $U \mapsto \nabla R(\bm{w};X)^TU$ with an appropriate matrix $U$. The algorithm for obtaining the inverse of the preconditioner is described in Algorithm \ref{algo:precon} below. 

\begin{table}[H]
{\LinesNumberedHidden
\begin{algorithm}[H] \label{algo:precon} 
\caption{Compute the inverse of preconditioner, $\bm{c} \in \W$}
\SetKwProg{fnc}{function}{}{end}
\SetAlgorithmName{Algorithm}{}

\fnc{$\mathrm{Precon}(\bm{w}, X)$}
{
$C_\ell \gets \zeros(m_{\ell}+1, m_{\ell+1}), \quad \ell = 1,\ldots,k$\;
\For{$i = 1,\ldots,n$}
{
$\bm{u}_i$ is sampled from $\mathrm{Unif}\left\{\pm 1\right\}^{m_1}$\;
$\bm{f}_{k} \gets \bm{u}_i \circ \bm{s}'_{k,i}$\;
\For{$\ell = (k-1),\ldots,1$}
{
$\bm{f}_\ell \gets W_{\ell+1}^{-}\bm{f}_{\ell+1} \circ \bm{s}'_{\ell,i}$\;
}
$C_\ell \gets C_\ell + \big(\widetilde{\bm{s}}_{\ell-1,i} \bm{f}_\ell^T \big)^2$,\quad $\ell=1,\ldots,k$\tcp*{Squares are taken componentwise}
}
$C_\ell \gets 1/[1 + \sqr(C_\ell/n)]$, \quad $\ell=1,\ldots,k$\tcp*{All operations are componentwise}
\KwRet $(C_1, C_2, \ldots, C_{k})$\;
}
\end{algorithm}}
\end{table}

\subsection{Initialization \& number of iterations in LSMR} \label{init_LSMR}
For sufficiently large $\lambda$, a modest number of iterations would suffice to produce an adequate approximate solution in LSMR. As $\lambda$ decreases, more iterations are required to account for pathological behaviors in curvature that may occur. To accommodate this, we increase the maximum number of iterations allowed in LSMR as HF iterations proceed. In order to avoid having too many parameters to tune, we update the number of LSMR iterations by
\[
\maxiter_{i+1} \gets \left\lceil\frac{n_{i+1}}{n_i} \cdot \maxiter_i \right\rceil, \quad i \geq 1,
\]
where $\maxiter_i$ represents the maximum number of LSMR iterations and $n_i$ denotes the mini-batch size at the $i$th HF iteration.

\vspace{2mm}\noindent
As for initializing LSMR, we use a simple yet clever idea proposed by \citet{martens2012training} and used by \citet{kiros2013training} to take advantage of sharing information across HF iterations. At iteration $i$, we denote the solution of LSMR as $\bm{d}_i$. The idea is to initialize LSMR at iteration $i$ from $\bm{d}_{i-1}$ with a decay factor $\gamma_i \in (0,1)$. Specifically, the initial solution to LSMR at iteration $i$ is $\gamma_i \bm{d}_{i-1}$ where the decay factor is updated by
\[
\gamma_{i+1} = \min(1.002 \cdot \gamma_{i},\ 0.95), \quad i \geq 1.
\]
Here we have a trade-off for the values of $\gamma_i$. At one end of the spectrum, when $\gamma_i=0$, we have the initial $\bm{d}_i = \bm{0}$. On the other end, when $\gamma_i=1$, we have the initial $\bm{d}_i = \bm{d}_{i-1}$. According to \citet{martens2012training}, the trade-off between $\bm{d}_{i-1}$ and $\bm{0}$ to be used to initialize LSMR works as follows: The previous iteration's solution $\bm{d}_{i-1}$ may be more converged than $\bm{0}$ along eigendirections that are more numerous or have small and spread-out eigenvalues. In contrast, $\bm{d}_{i-1}$ may be less converged than $\bm{0}$ along eigendirections that are fewer in number or have large and tightly clustered eigenvalues. The former group of eigendirections corresponds to low-curvature directions that tend to remain descent directions throughout many HF iterations. Since the latter directions are easier for LSMR to optimize than the former, it is preferable to initialize LSMR from the previous solution $\bm{d}_{i-1}$ over the zero vector $\bm{0}$, given that enough iterations are applied. 

\vspace{2mm}\noindent
In the early stages of HF optimization, we have aggressive LSMR truncations, which limit LSMR's ability to modify the update $\bm{d}$ from its initial value $\gamma_i \bm{d}_{i-1}$. To mitigate this, we start with a relatively small decay value, such as $\gamma_1 = 0.7$. In the later stages of the optimization, we utilize longer LSMR iterations. This enables us to increase the decay value $\gamma_i$ up to $0.95$, which helps to share more information across iterations towards the end of the training process. This, in turn, increases LSMR's ability to make significant progress along low-curvature directions. We stop increasing $\gamma_i$ at $0.95$ to correct the update direction $\bm{d}_{i-1}$ when the quadratic approximation varies across HF iterations.

\subsection{Computing the matrix-vector products} \label{matrix-vector_prods}
In the LSMR method, similar to other Krylov subspace methods, we utilize the system matrix A as an operator. This means that the matrix-vector products $A\bm{v}$ and $A^T\bm{u}$ can be computed through the mappings $\bm{v} \mapsto A\bm{v}$ and $\bm{u} \mapsto A^T \bm{u}$. In our notation, these mappings correspond to $\bm{d} \mapsto \nabla R(\bm{w}; X)\,\bm{d}$, known as the \textit{forward mode}, and $U \mapsto \nabla R(\bm{w};X)^TU$, known as the \textit{backward mode}.


\vspace{2mm}\noindent
In order to compute the products involving the derivatives of $R$, we need the network activations $S_\ell$ as defined in \eqref{S} as well as the derivatives of activations defined below:
\begin{equation} \label{Sprime}
S'_\ell \coloneqq 
\left[ \begin{array}{ccc}
\llongdash & (\bm{s}'_{\ell,1})^T & \rlongdash \\
\llongdash & (\bm{s}'_{\ell,2})^T & \rlongdash \\
 & \vdots & \\
\llongdash & (\bm{s}'_{\ell,n})^T & \rlongdash
\end{array} \right]
\coloneqq \sigma'_\ell\!\left( \widetilde{S}_{\ell-1}W_\ell \right)
\in \R^{n \times m_{\ell+1}}, \qquad \ell=1,2, \ldots,k.
\end{equation}
The quantities $S_\ell$ and $S'_\ell$ can be precomputed and cached since the weights $\bm{w}$ and data matrix $X$ remain fixed throughout the LSMR algorithm. This allows us to retrieve them for all the matrix-vector products made during the entire run of LSMR. Additionally, for the sake of brevity, we represent the weight matrices without the bias vectors as
\[
W_\ell^- \coloneqq W_\ell[1\!:\!m_\ell,\, 1\!:\!m_{\ell+1}] \in \R^{m_\ell \times m_{\ell+1}}, \qquad \ell=1,2,\ldots,k.
\]
The algorithms for computing the matrix-vector products are provided in Algorithm \ref{algo:forward} \& Algorithm \ref{algo:backward}.

\subsubsection{Forward mode}
For given vector $\bm{d} \in \mathcal{W}$, we compute $\nabla R(\bm{w};X)\,\bm{d}$ as follows:
\[
\nabla R(\bm{w}; X)\,\bm{d} = 
\left[
\begin{array}{cccc}
\nabla_1 R & \nabla_2 R & \cdots & \nabla_{k} R
\end{array}
\right]
\left[
\begin{array}{c}
D_1 \\ D_2 \\ \vdots \\ D_{k}
\end{array}
\right] =
\sum_{\ell=1}^{k} (\nabla_\ell R) \dd D_\ell
\in \mathcal{X},
\]
where $\nabla_{\ell} R$ stands for $\nabla_{W_\ell} R(\bm{x}; X)$. Now, we need to compute $4^{\text{th}}$-order tensor and matrix products $(\nabla_\ell R) \dd D_\ell$ for $\ell=1,\ldots,k$. 

\begin{table}[H]
{\LinesNumberedHidden
\begin{algorithm}[H] \label{algo:forward} 
\caption{The mapping $\bm{d} \mapsto \nabla R(\bm{w}; X)\,\bm{d}$ from $\W$ to $\X$}
\SetKwProg{fnc}{function}{}{end}
\SetKwInOut{Given}{Given}
\SetKwInOut{Input}{Input}
\SetAlgorithmName{Algorithm}{}

\Given{$\bm{w} \in \mathcal{W}$; $X \in \X$; $S_\ell$ and $S'_\ell$ for $\ell=1,\ldots,k$}
\fnc{$\mathrm{A}(\bm{d})$}
{
Write $\bm{d} = (D_1,\ldots,D_k)$\;
$F_1 \gets S'_1 \circ \widetilde{X}D_1$\;
\For{$\ell = 2,\ldots,k$}
{
$F_\ell \gets S'_\ell \circ \big( F_{\ell-1} W_{\ell}^- + \widetilde{S}_{\ell-1} D_\ell \big)$\;
}
\KwRet $\frac{1}{\sqrt{n}} F_{k}$\;
}
\end{algorithm}}
\end{table}

\subsubsection{Backward mode}
For given matrix $U \in \mathcal{X}$, we compute $\nabla R(\bm{w};X)^T U$ as follows:
\[
\nabla R(\bm{w}; X)^T U = 
\left[
\begin{array}{c}
(\nabla_1 R)^T \\
(\nabla_2 R)^T \\
\vdots \\
(\nabla_{k} R)^T
\end{array}
\right] U
=
\left[
\begin{array}{c}
(\nabla_1 R)^T \dd U \\
(\nabla_2 R)^T \dd U \\
\vdots \\
(\nabla_{k} R)^T \dd U 
\end{array}
\right] \in \mathcal{W},
\]
where $\nabla_{\ell} R \coloneqq \nabla_{W_\ell} R(\bm{x}; X)$.

\begin{table}[H]
{\LinesNumberedHidden
\begin{algorithm}[H] \label{algo:backward} 
\caption{The mapping $U \mapsto \nabla R(\bm{w};X)^T U$ from $\X$ to $\W$}
\SetKwProg{fnc}{function}{}{end}
\SetKwInOut{Given}{Given}
\SetKwInOut{Input}{Input}
\SetAlgorithmName{Algorithm}{}

\Given{$\bm{w} \in \mathcal{W}$; $X \in \mathcal{X}$; $S_\ell$ and $S'_\ell$ for $\ell=1,\ldots,k$}
\fnc{$\mathrm{A}^T(U)$}
{
$F_{k} \gets U \circ S'_{k}$\;
\For{$j = (k-1),\ldots,1$}
{
$F_\ell \gets F_{\ell+1} (W_{\ell+1}^-)^T \circ S'_\ell$\;
}
$D_{\ell} \gets \frac{1}{\sqrt{n}}\,\widetilde{S}_{\ell-1}^T F_{\ell}$, \quad $\ell=1,\ldots,k$\;
\KwRet $(D_1, D_2, \ldots, D_{k})$\;
}
\end{algorithm}}
\end{table}

\subsection{Applying LSMR in our notation} \label{LSMR_notation}
The LSMR method is thoroughly explained in Section \ref{LSMR}. It is important to note that we are using the same notation as \citet{fong2011lsmr} in Section \ref{LSMR}. Therefore, the variables defined in Section \ref{LSMR} are not linked to those defined earlier with the same letter.

\vspace{2mm}\noindent
In Section \ref{LSMR}, we solve the linear system $A\bm{x}=\bm{b}$ where $A$ is a matrix and $\bm{b}$ is a vector. As mentioned in Section \ref{autoencoders}, we do not vectorize the weight matrices $W_\ell$. Therefore, $A$ is a row vector of 4th-order tensors, and $\bm{b}$ is a matrix in our design. Note that Algorithm \ref{algo:LSMR} is compatible with these data types. In our design, the arguments of the $\mathrm{LSMR}$ function have the following names and types:
\[
\bm{x}_0 = \bm{d} \in \mathcal{W}, \quad 
A = \nabla R(\bm{w}; X) \in \mathcal{J}, \quad 
\bm{b} = -R(\bm{w}; X) \in \mathcal{X}, \quad
\lambda \in \R, \quad \text{and} \quad \bm{c} \in \mathcal{W}.
\]

\section{Least squares minimal residual (LSMR) method} \label{LSMR}
LSMR \citep{fong2011lsmr} is an iterative method for solving sparse least-squares problems 
\[
\min_{\bm{x} \in \R^n} \left\| A\bm{x}-\bm{b} \right\|_2^2, 
\]
where $A \in \R^{m \times n}$ and $\bm{b} \in \R^m$. The method is based on the Golub-Kahan bidiagonalization process. It is equivalent to the MINRES algorithm \citep{paige1975solution} applied to the normal equations
\begin{equation} \label{normal_eqn}
A^TA\,\bm{x} = A^T\bm{b}.
\end{equation}
Let $\bm{r}_k \coloneqq \bm{b} - A\bm{x}_k$ denote the residual for the approximate solution $\bm{x}_k$ in iteration $k$, and let $\bm{x}_0 \in \R^n$ be an initial solution. The approximate solution $\bm{x}_k$ is obtained by minimizing the residual of normal equations \eqref{normal_eqn} with respect to the $2$-norm over the affine space $\bm{x}_0 + \mathcal{K}_k(A^TA, A^T\bm{r}_0)$, where 
\[
\mathcal{K}_k(A^TA, A^T\bm{r}_0) \coloneqq \Span \left\{ A^T\bm{r}_0, (A^TA)A^T\bm{r}_0, \ldots, (A^TA)^{k-1}A^T\bm{r}_0 \right\}
\]
is a $k$-dimensional Krylov subspace. That is, the $k$th approximate solution $\bm{x}_k$ solves the problem
\begin{equation} \label{min_Ar}
\min_{\bm{x} \in \bm{x}_0 + \mathcal{K}_k(A^TA, A^T\bm{r}_0)} \left\| A^T(\bm{b}-A\bm{x}) \right\|_2.
\end{equation}
In LSMR, the quantities $\left\|\bm{r}_k\right\|_2$ and $\left\|A^T\bm{r}_k\right\|_2$ decrease monotonically, which allows for tractable early termination.

\vspace{2mm}\noindent
LSMR can be extended to the regularized least-squares problem
\begin{equation} \label{regLS}
\min_{\bm{x}} \left\| A\bm{x}-\bm{b} \right\|_2^2 + \lambda^2 \left\|\bm{x} \right\|_2^2 = \min_{\bm{x}} \left\| \left[ \begin{array}{c} A \\ \lambda I \end{array} \right]\bm{x} - \left[ \begin{array}{c} \bm{b} \\ \bm{0} \end{array} \right] \right\|_2^2, 
\end{equation}
where $\lambda$ is a regularization parameter. Define 
\begin{equation} \label{Abar}
\overline{A} \coloneqq \left[ \begin{array}{c} A \\ \lambda I \end{array} \right] \qquad \text{and} \qquad 
\overline{\bm{r}}_k \coloneqq \left[ \begin{array}{c} \bm{b} \\ \lambda\bm{x}_0 \end{array} \right] -\overline{A}\bm{x}_k
= \left[ \begin{array}{c} \bm{r}_k \\ \lambda(\bm{x}_0-\bm{x}_k) \end{array} \right].
\end{equation}
For inconsistent systems, the following stopping criterion given by \citet{fong2011lsmr} is effective:
\[
\big\|\overline{A}^T \overline{\bm{r}}_k \big\| \leq \mathrm{atol}\, \|\overline{A}\|\, \|\overline{\bm{r}}_k\|,
\]
where $\mathrm{atol}$ is a user-defined tolerance, and $\overline{A}$ \& $\overline{\bm{r}}_k$ are counterparts of $A$ \& $\bm{r}_k$ for regularized least-squares problems and defined in \eqref{Abar}. In addition to the criterion above, we monitor the progress in some merit function $\phi(\bm{x}_k)$ over the iterations $\bm{x}_k$. We choose the objective function for validation dataset as a merit function:
\begin{equation} \label{merit}
\phi(\bm{d}) \coloneqq f(\bm{w}+\bm{d}; X_{\val}).
\end{equation}
We plot the objective values for the validation datasets to trace the progress over LSMR iterations. There are two prevalent behaviors observed: either (i) the objective value first decreases rapidly and then stagnates, or (ii) it decreases and then increases and then decreases again. To address the first type of behavior, we terminate the iterations once the relative progress made in $\phi(\bm{x})$ per-iteration fell below some tolerance $\mathrm{ftol}$. 

\vspace{2mm}\noindent
Since evaluating $\phi$ at each iterate $\bm{x}_k$ would be computationally expensive, we compute the relative progress in $\phi$ per-iteration over progressively larger intervals as $k$ grows. Averaging over multiple iterations also helps us obtain a more reliable estimate, since there is a high degree of variability in the relative reduction in $\phi$. Starting with $k_{\mathrm{prev}} = 0$, $k_{\mathrm{next}}=5$, we compute relative progress over the interval $(k_{\mathrm{prev}}, k_{\mathrm{next}})$. We follow \citeauthor{martens2010deep}'s practice and update $k_{\mathrm{next}} \gets \lceil 1.25 \cdot k \rceil$ and $k_{\mathrm{prev}} \gets k$ whenever $k = k_{\mathrm{next}}$. In particular, we terminate LSMR at iteration $k$ if the following condition is satisfied:
\[
k_{\mathrm{next}} = k > \mathrm{miniter} \qquad \text{and} \qquad
f_k = f_{\min} \qquad \text{and} \qquad
\frac{f_{\mathrm{prev}} - f_k}{f_k} < (k-k_{\mathrm{prev}})\,\mathrm{ftol},
\]
where $\mathrm{miniter}$ is the minimum number of iterations\footnote{$\mathrm{miniter}$ is the minimum number of iterations as long as $\|\overline{A}^T \overline{\bm{r}}_k \| > \mathrm{atol}\, \|\overline{A}\|\, \|\overline{\bm{r}}_k \|$. Whenever $\|\overline{A}^T \overline{\bm{r}}_k \| \leq \mathrm{atol}\, \|\overline{A}\|\, \|\overline{\bm{r}}_k \|$, the LSMR iterations stop even though $k < \mathrm{miniter}$.}, 
$f_k \coloneqq \phi(\bm{x}_k)$ and $f_{\mathrm{prev}} \coloneqq \phi(\bm{x}_{k_{\mathrm{prev}}})$ are the values of merit function at iteration $k$ and $k_{\mathrm{prev}}$ respectively, and 
$f_{\min}$ is the minimum value of $\phi(\bm{x}_\ell)$ over the iterations $\ell=1,\ldots,k$, 

\begin{table}[H]
{\LinesNumberedHidden
\begin{algorithm}[H] \label{algo:LSMR} 
\caption{Preconditioned LSMR}
\SetKwProg{fnc}{function}{}{end}
\SetKwInOut{Given}{Given}
\SetAlgorithmName{Algorithm}{}

\fnc{$\mathrm{LSMR}(\bm{x}_0, A, \bm{b}, \lambda, \bm{c}; \maxiter, \atol, \ftol, \phi)$}
{
$\bm{r}_0 \gets \bm{b} - A(\bm{c} \circ \bm{x}_0)$; \quad
$\beta_1 \bm{u}_1 \gets \bm{r}_0$; \quad 
$\alpha_1 \bm{v}_1 \gets \bm{c} \circ A^T \bm{u}_1$; \quad
$\overline{\alpha}_1 \gets \alpha_1$; \quad
$\overline{\zeta}_1 \gets \alpha_1\beta_1$\;
$\rho_0 \gets 0$; \quad
$\overline{\rho}_0 \gets 1$; \quad
$\overline{c}_0 \gets 1$; \quad
$\overline{s}_0 \gets 0$; \quad 
$\bm{h}_1 \gets \bm{v}_1$; \quad
$\overline{\bm{h}}_0 \gets \bm{0}$\;
$f_{\min} \gets \infty$; \quad 
$f_{\mathrm{prev}} \gets \phi(\bm{x}_0)$; \quad
$k_{\min} \gets 0$; \quad
$k_{\mathrm{prev}} \gets 0$; \quad
$k_{\mathrm{next}} \gets 5$; \quad
$\bm{x}_{\min} \gets \bm{x}_0$\;
$\mathrm{miniter} \gets 50$; \quad
$\mathrm{recover} \gets 100$\;
\For{$k = 1,\ldots,\mathrm{maxiter}$}{
\tcp{Golub-Kahan bidiagonalization}
$\beta_{k+1} \bm{u}_{k+1} \gets A(\bm{c} \circ \bm{v}_k) - \alpha_k \bm{u}_k$\;
$\alpha_{k+1} \bm{v}_{k+1} \gets \bm{c} \circ A^T \bm{u}_{k+1} - \beta_{k+1} \bm{v}_k$\;
\tcp{Construct rotation $\widehat{P}_k^{(2k+1)}$}
$\widehat{\alpha}_k \gets (\overline{\alpha}_k^2 + \lambda^2)^{1/2}$; \quad
$\widehat{c}_k \gets \overline{\alpha}_k/\widehat{\alpha}_k$; \quad
$\widehat{s}_k \gets \lambda/\widehat{\alpha}_k$\;
\tcp{Construct and apply rotation $P_k^{(2k+1)}$}
$\rho_k \gets (\widehat{\alpha}_k^2 + \beta_{k+1}^2)^{1/2}$; \quad
$c_k \gets \widehat{\alpha}_k/\rho_k$; \quad
$s_k \gets \beta_{k+1}/\rho_k$; \quad
$\theta_{k+1} \gets s_k \alpha_{k+1}$; \ \ \
$\overline{\alpha}_{k+1} \gets c_k \alpha_{k+1}$\;
\tcp{Construct and apply rotation $\overline{P}_k^{(2k+1)}$}
$\overline{\theta}_k \gets \overline{s}_{k-1} \rho_k$; \quad
$\overline{\rho}_k \gets \left( \overline{c}_{k-1}^2\rho_k^2 + \theta_{k+1}^2 \right)^{1/2}$; \quad
$\overline{c}_k \gets \overline{c}_{k-1} \rho_k/\overline{\rho}_k$; \quad
$\overline{s}_k \gets \theta_{k+1}/\overline{\rho}_k$\;
$\zeta_k \gets \overline{c}_k \overline{\zeta}_k$; \quad
$\overline{\zeta}_{k+1} \gets -\overline{s}_k \overline{\zeta}_k$\;
\tcp{Update $\overline{\bm{h}}_k, \bm{x}_k, \bm{h}_k$}
$\overline{\bm{h}}_k \gets \bm{h}_k - \dfrac{\overline{\theta}_k \rho_k}{\rho_{k-1}\overline{\rho}_{k-1}} \overline{\bm{h}}_{k-1}$; \quad \
$\bm{x}_k \gets \bm{x}_{k-1} + \dfrac{\zeta_k}{\rho_k \overline{\rho}_k} \overline{\bm{h}}_k$; \quad \
$\bm{h}_{k+1} \gets \bm{v}_{k+1} - \dfrac{\theta_{k+1}}{\rho_k} \bm{h}_k$\;
\tcp{Termination conditions}
\textbf{if} $\big\|\overline{A}^T \overline{\bm{r}}_k \big\| \leq \mathrm{atol}\, \|\overline{A}\|\, \|\overline{\bm{r}}_k \|$ \textbf{then break}\tcp*{$\overline{A}$ and $\overline{\bm{r}}_k$ are defined in \eqref{Abar}}
\If{$k = k_{\mathrm{next}}$}
{
$f_{k} \gets \phi(\bm{x}_k)$\tcp*{$\phi$ is a merit function defined in \eqref{merit}}
\lIf{$f_k < f_{\min}$}
	{
	$f_{\min} \gets f_k$; \quad
	$k_{\min} \gets k$; \quad
	$\bm{x}_{\min} \gets \bm{x}_k$
	}
\lIf{$(k > \mathrm{miniter})$  $\bm{\mathrm{and}}$ $(f_k = f_{\min})$ $\bm{\mathrm{and}}$ $\dfrac{f_{\mathrm{prev}} - f_{\val}}{f_{\val}} < (k-k_{\mathrm{prev}})\,\mathrm{ftol}$}{\textbf{break}}
\lIf{$(k > \mathrm{miniter})$  $\bm{\mathrm{and}}$ $(f_k > f_{\min})$ $\bm{\mathrm{and}}$ $k > k_{\min} + \mathrm{recover}$}{\textbf{break}}
$k_{\mathrm{next}} \gets \min\!\left( \lceil 1.25 \cdot k_{\mathrm{next}} \rceil, \mathrm{maxiter} \right)$; \quad 
$k_{\mathrm{prev}} \gets k$; \quad
$f_{\mathrm{prev}} \gets f_k$\;
}
}
\KwRet $\bm{x}_k$
}
\end{algorithm}}
\end{table}
\noindent
To address the second type of behavior, we wait for $\phi(\bm{x}_k)$ to recover for a certain number of iterations, denoted by $\mathrm{recover}$. In our experiments, this typically takes no more than $20$ iterations. We terminate LSMR at iteration $k$ if it has not recovered in $\mathrm{recover}$ iterations after $k_{\min}$ as shown below:
\[
k_{\mathrm{next}} = k > \mathrm{miniter} \qquad \text{and} \qquad
f_k > f_{\min} \qquad \text{and} \qquad
k > k_{\min} + \mathrm{recover}. 
\]
Algorithm \ref{algo:LSMR} outlines how to use LSMR to solve the regularized least-squares problem \eqref{regLS} with given initial solution $\bm{x}_0 \in \R^n$, regularization parameter $\lambda \geq 0$, and preconditioner $\bm{c} \in \R^n$. Note that $\mathrm{maxiter}$ denotes the maximum number of iterations.

\section{Experiments} \label{experiments}
We conduct a series of experiments to evaluate the effectiveness of our SHF approach on deep autoencoders. In these experiments, we focus on both training and test errors. Training errors help us understand the effectiveness of our optimization method, while test errors allow us to gauge how well the model can generalize to new data. We also compare the performance of our algorithm with the HF optimization algorithm proposed by \citet{martens2010deep}. 

\vspace{2mm}\noindent
We intended to use the same datasets (CURVES, MNIST, and FACES) as \citet{martens2010deep} and \citet{hinton2006reducing}. However, the authors did not provide the preprocessed data or preprocessing code for the augmented Olivetti face dataset, FACES. For this reason, we conduct the experiments using the CURVES, MNIST, and USPS datasets, which are explained below:
\begin{itemize}
\item[(a)] The CURVES dataset is a \textit{synthetic} dataset containing images of curves generated from three randomly chosen points in two dimensions by \citet{hinton2006reducing}. Each curve is a $28 \times 28$ pixel grey-scale image. The dataset contains 30,000 images and is available on \citeauthor{martens2010deep}' website at \href{http://www.cs.toronto.edu/~jmartens/digs3pts_1.mat}{http://www.cs.toronto.edu/$\sim$jmartens/digs3pts\_1.mat}.

\item[(b)] The MNIST database is a widely used dataset of hand-written digits. Each digit is represented as a $28 \times 28$ pixel grey-scale image. The dataset consists of 60,000 training images and 10,000 test images. The dataset is available on Yann LeCun's website at \href{http://yann.lecun.com/exdb/mnist/}{http://yann.lecun.com/exdb/mnist/}.

\item[(c)] The USPS dataset consists of hand-written digits, each represented by a $16 \times 16$ pixel grey-scale image. There are 11,000 images in total, with 1100 examples of each digit. This dataset can be found on Sam Roweis' website at \href{https://cs.nyu.edu/~roweis/data/usps_all.mat}{https://cs.nyu.edu/$\sim$roweis/data/usps\_all.mat}.
\end{itemize}
In each dataset, the pixel values are normalized to the range $[0,1]$, and the original $28 \times 28$ (or $16 \times 16$) matrices are flattened to vectors of dimension $784$ (or $256$). In our experiments, logistic activation functions are used in each layer of autoencoders. It is worth noting that the logistic activations outperform the ReLU activations in our experiments.  

\vspace*{-\baselineskip}
\addtolength{\tabcolsep}{3pt}
\begin{table}[H]
\centering
\caption{Experimental parameters}
\begin{tabular}{c c c c c}
\toprule
\textbf{Dataset} & \textbf{Train - test - val sizes} & $ n_1$ & $n_{\max}$ & \textbf{Encoder dimensions}  \\ \midrule
CURVES	& 20000 - 8000 - 2000 	& 250 & 2500 &  784 - 400 - 200 - 100 - 50 - 25 - 6 \\
MNIST	& 50000 - 10000 - 10000	& 300 & 6000 &  784 - 1000 - 500 - 250 - 30 \\
USPS	& 8000 - 2000 - 1000  & 200	& 2000 &  256 - 400 - 200 - 100 - 50 - 25 \\ \bottomrule
\end{tabular}
\label{table:params} 
\end{table}

\noindent
In Table \ref{table:params}, you can find the experimental parameters for each dataset. The table includes the sizes of the training, test, and validation data in the second column. Additionally, it shows the initial mini-batch size ($n_1$), the maximum mini-batch size allowed ($n_{\max}$), and the encoder dimensions in the last column. Note that we consider a symmetric autoencoder, where the decoder architecture is the mirror image of the encoder.

\vspace{1mm}\noindent
Following the principles of reproducible research, the source code and data files necessary to replicate the experimental results of this paper can be accessed from \href{https://github.com/emirahmet-ibrahim}{https://github.com/emirahmet-ibrahim}. For the sake of completeness, we also provide the hyper-parameters used in the experiments for the CURVES, MNIST, and USPS datasets, respectively:
\begin{itemize}[leftmargin=6mm]
\setlength\itemsep{-0.1em}
\item[(a)] $\lambda_1 = 5, \drop = \frac{99}{100}, \gamma_1 = 0.7, \maxiter_1 = 200, \atol=10^{-8}, \ftol = 10^{-7}, \theta=0.5, m_0=20$, 
\item[(b)] $\lambda_1 = 12, \drop = \frac{49}{50}, \gamma_1 = 0.7, \maxiter_1 = 150, \atol=10^{-8}, \ftol = 10^{-5}, \theta=0.2, m_0=10$, 
\item[(c)] $\lambda_1 = 7.5, \drop = \frac{99}{100}, \gamma_1 = 0.65, \maxiter_1 = 150, \atol=10^{-6}, \ftol = 2\cdot 10^{-5}, \theta=0.2, m_0=10$. 
\end{itemize}
Moreover, we set $\sigma = 1.5$ and $\alpha = 10^{-4}$ in Algorithm \ref{main} for all datasets. The most important hyper-parameters are $\lambda_1, \drop, \maxiter_1$, and $\ftol$. A good choice of the parameter $\lambda_1$ depends closely on the dataset at hand. The parameter $\drop$ can be chosen from the range $[\frac{2}{3}, \frac{99}{100}]$. The higher values usually yields a better generalization (i.e. lower test errors). A good range for the parameter $\maxiter_1$ is $[150, 250]$. The parameter $\ftol$ is also highly dependent to the dataset. In general, the higher values of $\ftol$ leads to the lower test errors. 

\vspace{1mm}\noindent
We implemented our method in Julia v1.10.3. We also rewrote \citet{martens2010deep}'s HF algorithm in Julia using his provided code as a basis and re-ran his experiments over the datasets mentioned above. We managed to get approximately the same results as he reported for both the CURVES and MNIST experiments. Note that the USPS dataset is not used in \citet{martens2010deep}.

\vspace{1mm}\noindent
In order to get a more accurate assessment of how well these algorithms perform, we ran each algorithm five times with the same hyper-parameters but different random seeds, and then average the results. Table \ref{table:results} shows the overall experimental results. For each run, we recorded the errors produced by the algorithms as follows: for our algorithm, we stopped once we achieve the lowest validation error, so the training and test errors are those corresponding to the lowest validation error for each run. For the algorithm by \citet{martens2010deep}, we simply selected the best test error and its corresponding training error from each run.

\vspace*{-\baselineskip}
\begin{table}[H]
\centering
\caption{Experimental results}
\begin{tabular}{lcccc}
\toprule
\multirow{2}{*}{Dataset} &  \multicolumn{2}{c}{\citet{martens2010deep}} & \multicolumn{2}{c}{Our method}  \\ \cmidrule(r){2-3} \cmidrule{4-5}
& Training error &  Test error & Training error & Test error \\ \midrule
CURVES	& 0.051 & 0.098 & 0.160 & 0.214  \\
MNIST	& 0.910 & 1.436 & 0.959 & 1.435  \\
USPS	& 1.007 & 1.961 & 0.999 & 1.947  \\ \bottomrule
\end{tabular}
\label{table:results} 
\end{table}

\begin{figure}[H]
\centering
\includegraphics[scale=0.5]{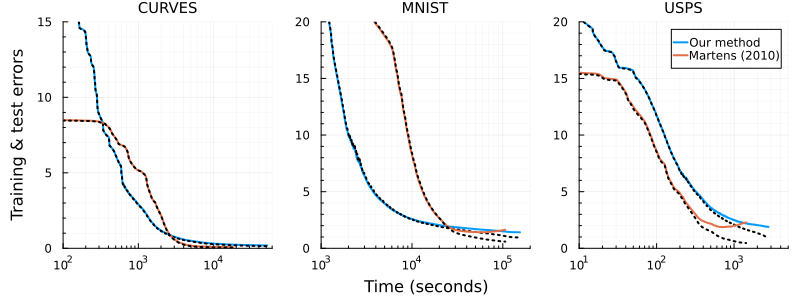}
\caption{Training (dotted) and test (solid) errors vs. computation time}
\label{fig:errors}
\end{figure}

\begin{figure}[H]
\centering
\includegraphics[scale=0.53]{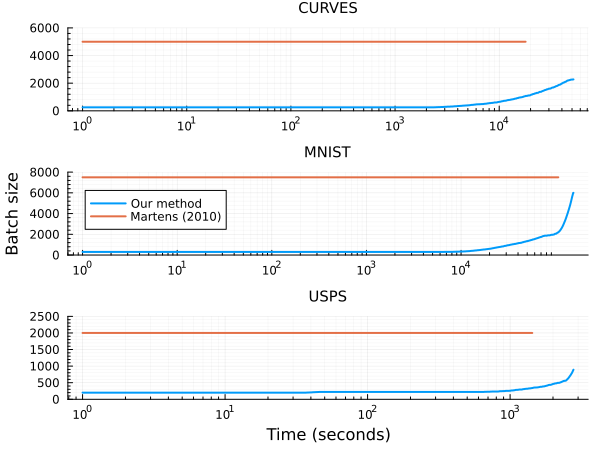}
\caption{Mini-batch size vs. computation time}
\label{fig:batch}
\end{figure}

\noindent
We also provide the training and test error curves over time for both methods in Figure \ref{fig:errors}. The dotted and solid lines correspond to the training and test curves, respectively. The curves produced by our method follow similar path as in \citet{martens2010deep}, indicating the efficiency of our approach. Figure \ref{fig:batch} demonstrates the evolution of mini-batch sizes over computation time in HF optimization. \citet{martens2010deep} uses a fixed mini-batch size, partitioning the training data into batches of constant size and rotating through them. As detailed in Section \ref{mini-batch_size}, we gradually increase the mini-batch size to save time and memory. Overall, our approach uses significantly less memory during training, as illustrated in Figure \ref{fig:batch}. 

\noindent
The experimental results show that our method performs similarly to \citet{martens2010deep} on these datasets. As shown in Table \ref{table:results}, we achieve slightly better test errors for the MNIST and USPS datasets, while \citet{martens2010deep} outperforms our method in the synthetic CURVES dataset. Our algorithm is designed to prevent overfitting, which explains why it performs better on real-world datasets. Overall, these results are quite promising. Our SHF optimization algorithm demonstrates strong test accuracy without compromising training speed.

\section{Conclusion} \label{conclusion}
This paper presents a novel approach to accelerate Hessian-free (HF) optimization for training deep autoencoders by leveraging the LSMR method and an improved mini-batch selection strategy. Our approach enhances the efficiency of the HF algorithm by addressing the challenges associated with large sparse linear systems and mitigating overfitting through dynamic mini-batch sizing. The integration of LSMR for estimating update directions, combined with \citet{chapelle2011improved}'s improved preconditioner and our new mini-batch selection algorithm, has demonstrated significant improvements in both reducing data usage and generalization performance. Experimental results confirm that our stochastic HF optimization method enables rapid training while maintaining better generalization error compared to traditional approaches. This work not only enhances the optimization of deep autoencoders, but also establishes a foundation for further progress in Hessian-free methods and their application in large-scale machine learning tasks.

\bibliography{../figures/mybib}

\end{document}